\documentclass[journal]{IEEEtran}

\usepackage{cite}
\ifCLASSINFOpdf
   \usepackage[pdftex]{graphicx}
\else
   \usepackage[dvips]{graphicx}
\fi
\usepackage{amsmath}
\ifCLASSOPTIONcompsoc
  \usepackage[subrefformat=parens, labelformat=parens,  caption=false, font=normalsize, labelfont=sf, textfont=sf]{subfig}
\else
  \usepackage[subrefformat=parens, labelformat=parens,  caption=false, font=footnotesize]{subfig}
\fi

\usepackage{cases}
\usepackage{siunitx}
\usepackage{pgfplots}
\usepgfplotslibrary{colorbrewer}
\usepackage{physics}
\pgfplotsset{compat=1.15} 
\usepackage{filecontents}
\usepackage{gensymb}
\usepackage[export]{adjustbox}

\usepackage{color,soul}
\usepackage{xcolor}

\usepackage[framemethod=tikz]{mdframed}

\begin{document}
\bstctlcite{IEEEexample:BSTcontrol}

\title{Design and Modeling of a Smart Torque-Adjustable Rotary Electroadhesive Clutch for Application in Human-Robot Interaction}

\author{Navid~Feizi,~\IEEEmembership{Student Member,~IEEE,}
	S.~Farokh~Atashzar,~\IEEEmembership{Member,~IEEE,}
	Mehrdad R.~Kermani,~\IEEEmembership{Member,~IEEE,}
	and~Rajni~V.~Patel,~\IEEEmembership{Life~Fellow,~IEEE}
	\thanks{The work of NF and RVP was funded by the Natural Sciences and Engineering Research Council (NSERC) of Canada under grant \#RGPIN-1345 (awarded to RVP) and the Tier-1 Canada Research Chairs Program (RVP). The work of MRK was funded by Natural Sciences and Engineering Research Council (NSERC) under the grant \#RGPIN-06253. The work of SFA was supported by the National Science Foundation (Award \#2037878).}
	\thanks{N.~Feizi (corresponding author) is with Canadian Surgical Technologies and Advanced
		Robotics (CSTAR), Lawson Health Research Institute, London, ON N6A 5A5,
		Canada, and with the School of Biomedical Engineering, Western University, London, ON N6A 3K7, Canada
		(email:~nfeizi@uwo.ca)}
	\thanks{S.~F.~Atashzar is with the Department of Mechanical and Aerospace Engineering and the Department of Electrical and Computer Engineering, and the Department of Biomedical Engineering, New York University (NYU), New York, NY 10003 USA. He is also with NYU WIRELESS and NYU CUSP.		
		(email:~f.atashzar@nyu.edu)}
	\thanks{M.~R.~Kermani is with the Department of Electrical and Computer Engineering, Western University, London,
		ON N6A 5B9, Canada
		(email: mkermani@eng.uwo.ca)}
	\thanks{R.~V.~Patel is with Canadian Surgical Technologies and Advanced
		Robotics (CSTAR), Lawson Health Research Institute, London, ON N6A 5A5,
		Canada, and with the Department of Electrical and Computer Engineering, the School of Biomedical Engineering, the Department of Surgery, and the Department of Clinical Neurological Sciences, Western University, London,
		ON N6A 5B9, Canada
		(email:~rvpatel@uwo.ca)}}

%

\IEEEpubid{978-1-5386-5541-2/18/\$31.00 ©2018 IEEE\copyright~2018 IEEE  -  The manuscript is under review in IEEE TMECH}


\maketitle

\begin{abstract}
	The increasing need for sharing workspace and interactive physical tasks between robots and humans has raised concerns regarding safety of such operations. In this regard, controllable clutches have shown great potential for addressing important safety concerns at the hardware level by separating the high-impedance actuator from the end effector by providing the power transfer from electromagnetic source to the human. However, the existing clutches suffer from high power consumption and large-weight, which make them undesirable from the design point of view. In this paper, for the first time, the design and development of a novel, lightweight, and low-power torque-adjustable rotary clutch using electroadhesive materials is presented. The performance of three different pairs of clutch plates is investigated in the context of the smoothness and quality of output torque. The performance degradation issue due to the polarization of the insulator is addressed through the utilization of an alternating current waveform activation signal. Moreover, the effect of the activation frequency on the output torque and power consumption of the clutch is investigated. Finally, a time-dependent model for the output torque of the clutch is presented, and the performance of the clutch was evaluated through experiments, including physical human-robot interaction. The proposed clutch offers a torque to power consumption ratio that is six times better than commercial magnetic particle clutches. The proposed clutch presents great potential for developing safe, lightweight, and low-power physical human-robot interaction systems, such as exoskeletons and robotic walkers.
\end{abstract}

\begin{IEEEkeywords}
Electroadhesion, smart actuators, adjustable clutch, rotary clutch.
\end{IEEEkeywords}

\IEEEpeerreviewmaketitle

\section{Introduction}
	\IEEEPARstart{W}{ith} the development of interactive robotics that shares workspace and tasks with humans, such as exoskeletons and rehabilitation robots, the safety of robotic systems has become an important issue \cite{zacharaki2020safety}. Controllers that consider the safety of the robotic system along with the transparency of interaction such as passivity controller \mbox{\cite{atashzar2016passivity, feizi2021time, feizi2022adaptive}} and wave variable controller \mbox{\cite{aziminejad2008transparent}} are two of the known methods that provide safety at the system control level. By designing the system based on passive components, such as clutches, safety can be assured at the mechanical level. Clutches have been used in several applications in robotics, particularly in physical human-robot interaction (pHRI) systems \mbox{\cite{lauzier2011series, khazoom2019design, diller2016lightweight}}. A clutch provides a limit on the torque/force coupling between the human and the source of mechanical energy leading to the intrinsic safety of the system. Clutches provide safety by separating the high-impedance actuator from the end effector, providing HRI safety in case of collisions or needed physical contacts for task-sharing \mbox{\cite{roveda2019assisting, shafer2010feasibility}}. Constant damping clutches \cite{mehrabi2019design} and on/off electromagnetic actuated clutches \cite{rouse2014clutchable} are the most common types used in the literature mainly for emergency scenarios.
	
	\IEEEpubidadjcol
	
	Using smart material and intelligent mechanisms, new types of clutches and actuators have been developed. Adjustable clutches allow a tunable amount of torque to be transferred thereby; they can be either used as part of the control loop of the  actuator \cite{hirata2006motion}, or combined with passive springs as energy-storing actuators \cite{diller2016lightweight}, or combined with motors (in parallel or series) as hybrid actuators \cite{dills2020hybrid, najmaei2015design, najmaei2014application}.
	
	Magnetic particle (MP) and magneto-rheological (MR) clutches are the two types of reputable torque adjustable clutches that work based on dry and viscous friction, respectively \cite{najmaei2015design, najmaei2014application, shafer2011design, pisetskiy2021high, PLACIDindustries2021}. In MP clutches, the electromagnetic field generated by the coil that is integrated into the clutch affects the formation of the ferromagnetic particles. In MR clutches the viscosity of the ferromagnetic fluid provides a means of tuning the torque. Thus, the damping of clutches can be adjusted by tuning the coil current. The integration of the ferromagnetic parts, the magnetic fluid, and a coil in the clutch structure results in relatively heavy-weight not desirable for several practical applications, especially in the context of wearable robotics such as exoskeletons \mbox{\cite{roveda2020design}}. In addition, the generation of the magnetic field leads to high power consumption, which again is a challenge for wearable and mobile applications \cite{moghani2016design, moghani2019lightweight, Wellborn}. In an effort to address these issues by eliminating heavy ferromagnetic parts and the electric coil, electro-rheological (ER) clutches have been considered as an alternative \cite{nikitczuk2009active, davidson2018electrorheological, boku2010development}. However, the low viscosity of ER fluids significantly reduces the performance of ER clutches in comparison to MR clutches.
	
	The electrostatic force has been harnessed to generate Electroadhesion (EA), mostly for object handling and robotic locomotion applications \cite{schaler2017electrostatic, shintake2018soft}. In contrast to the magnetic field, the required energy to generate an electric field is very small, and there is no need for heavy parts and fluid (which may have a leakage problem). The electrostatic force, therefore, shows high potential to be exploited for producing adjustable smart clutches, in a low-space, lightweight, and energy-efficient manner. In this regard, the EA force has been used as a binary joint locking mechanism for a robotic hand \cite{aukes2014design}. Diller et al. utilized EA force in a translational locking mechanism for a passive energy-storing ankle exoskeleton \cite{diller2016lightweight}. EA clutches have been used in haptic feedback-based wearable devices in \cite{ramachandran2019all, hinchet2020high}. Zhang et al. used EA clutch in a locking mechanism for a 2.5D tactile display \cite{zhang2019design}.  
	
	The EA force has not been limited to the above-mentioned applications and has also been used in applications such as active connection mechanisms, stiffness tuning in composite structures, and controllable perching mechanisms \cite{guo2019electroadhesion}. Despite the functionality in many applications, shear EA clutches have only been used to create binary static friction in locking mechanisms in an on/off mode. The lack of implementation of a device with adjustable EA shear force is perhaps caused by the degradation of the EA force with time due to residual charges, resulting in time-dependent dynamics of the EA force, which makes such an implementation complex \cite{aukes2014design, hinchet2020high, zhang2019design, chen2020time, graule2016perching}. To the best of our knowledge, all existing EA clutches (except \mbox{\cite{aukes2014design}} which was designed for locking a limited revolute joint), have been implemented with linear motion, which creates restrictions for most applications that require infinite rotations i.e., a walker. While linear EA clutches have been used in ankle exoskeletons \mbox{\cite{diller2016lightweight}}, a rotary EA clutch can simplify the mechanical design of the device and minimize space.
	
	In this paper, for the first time, the EA force has been harnessed in a rotary and tunable fashion to make a novel, lightweight, low-power, and torque adjustable rotational clutch. The contributions of the research are as follows:
	\begin{itemize}
		\item Design and development of a novel rotary clutch for harnessing and controlling the shear friction based on electroadhesion.
		\item Proposing and experimentally validating a novel actuation scheme based on an alternating current activation signal for eliminating torque degradation in EA clutches, and study the effect of frequency and activation voltage.
		\item Study the effect of material and surface finishes on the performance of EA clutch for output torque generation.
		\item Development and experimental validation of a general model for the EA clutch torque.
	\end{itemize}

	The remainder of this document is organized as follows. Section {\ref{sec: Concepts}} briefly explains the concept of EA clutch, which is built based on electrostatic force and the theory behind it. Section {\ref{sec: clutch design}} presents the mechanical design of the clutch, the fabrication procedure of every pair of clutch discs, and the design of the driver circuit. Section {\ref{sec: clutch discs selection}} compares the primary results of the three pairs of clutch discs and explains the reasons behind the selected pair. Section {\ref{sec: performance  eval and modeling}} evaluates the performance of the selected pair of clutch discs under two different activation regimes. A nonlinear model for the clutch torque is presented for each activation regime. Section {\ref{sec: HRI experiment}} presents the experimental validation of the clutch in pHRI. Section {\ref{sec: conclusions}} provides concluding remarks.
	
\section{Concepts}
\label{sec: Concepts}
	An electroadhesive force is an electrostatic force between two charged particles, whether they are conductors or insulators. However, the mechanism for EA force generation in conductors is different from that in insulators. The EA force in conductors is mainly generated from the attraction of charges on the electrodes. The shear stress on the interface of the electrodes is governed by Columb's law \cite{monkman1997analysis} as shown below:
	\begin{equation}	
		\sigma_{sh} = \dfrac{\epsilon_r \epsilon_0 V^2 C_f}{2d^2} 
		\label{eq: EA shear stress}
	\end{equation}
	where $\epsilon_r$ is the relative permittivity of the insulator, $\epsilon_0$ is the permittivity of vacuum, $V$ is the applied voltage across the electrodes, $C_f$ is the friction coefficient, and $d$ is the distance between the electrodes.
	
	Electroadhesive force in insulators results from the polarization of the dielectric, which cannot be formulated in detail without knowledge of the molecular structure of the dielectric. The electric field polarizes the insulator and results in an adhesion force that can be formulated \cite{monkman1997analysis} as
	\begin{equation}
		\sigma_{sh} = \dfrac{P E C_f}{A} 
		\label{eq: EA shear stress isolator}
	\end{equation}
	where $E$ is the electric field, $A$ is the overlap area, and $P$ is the polarization which can be simply formulated as follow for materials such as barium titanate,
	\begin{equation}
		P = P_o + P_s  
		\label{eq: polarization}
	\end{equation}
	where $P_o$ is the orientational polarization, and $P_s$ is the space charge polarization due to hopping and interfacial polarizations \cite{guo2016optimization}. The latter polarization has sluggish dynamics with a relatively large time constant.
	
	Based on the mechanical structure of the system, one or both of the above-mentioned mechanisms may be involved in the electroadhesion shear stress.

\section{Clutch Design and Fabrication}
\label{sec: clutch design}
	\subsection{Mechanical design}
\begin{figure}[!t]
			\centering
			\includegraphics[width=.9\linewidth,keepaspectratio,]{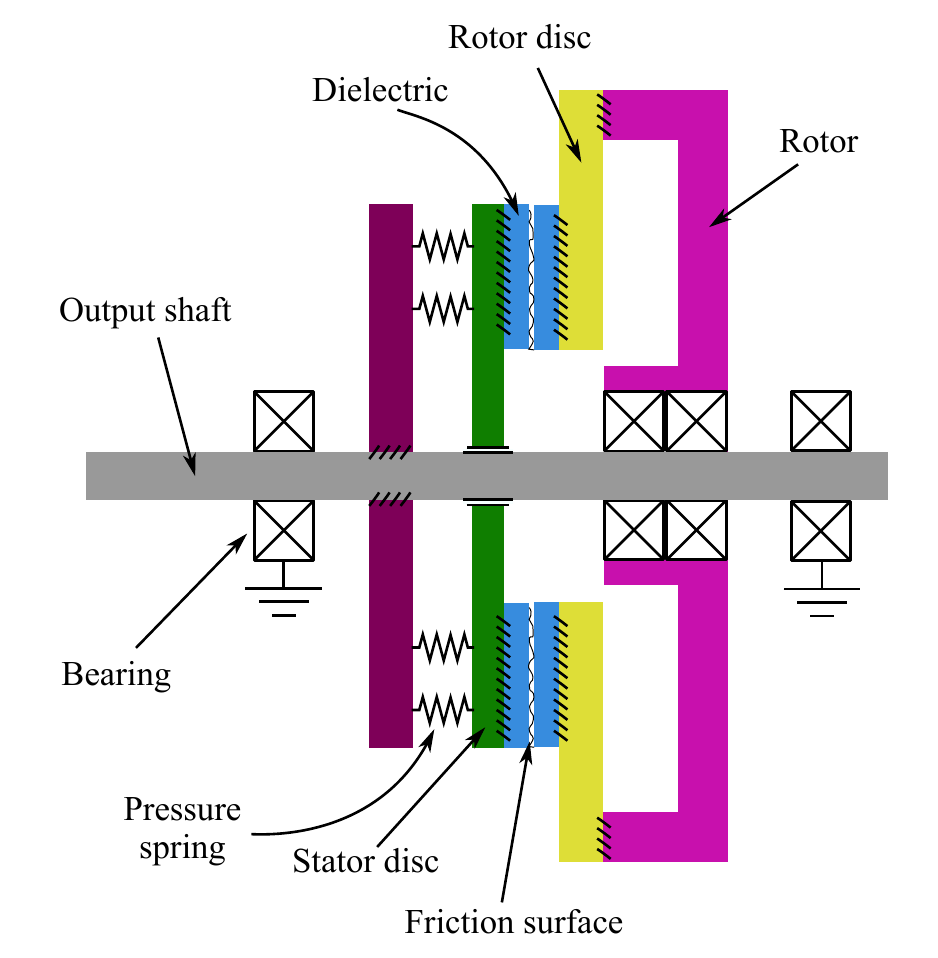}
			\caption{Schematic of the rotary electroadhesive clutch}
			\label{fig:  clutch schematic}
		\end{figure}

		The core of the proposed rotary EA clutch consists of a rotor disc (outer disc) and a stator disc (inner disc) sliding against each other. Each disc acts as an electrode; thus, the clutch can be analogously considered a capacitor. The applied voltage across the clutch discs generates an electric field that develops an attraction force between both conductors and insulators, as explained in Section \ref{sec: Concepts}. This creates a ring-like friction area that transfers the torque between the discs. The friction surfaces of the discs are covered with Dupont 8153 dielectric (Luxprint, Dupont Microcircuit Materials, NC, USA), which is a barium titanate-based dielectric to prevent electric discharge. Ideally, a small gap should be maintained between the discs to avoid undesirable friction and wear on the discs. At the same time, to maximize the EA forces and the response time of the clutch, a zero-gap should be kept between the discs. Addressing these two competing design requirements without adding other components and increasing the complexity and cost of the overall design is not possible. To this effect and for the purpose of this initial study, pressure springs were integrated into the design to continuously push the stator disc against the rotor disc with an infinitesimal air gap between the discs. The springs keep the discs ready to engage and reduce the activation time of the clutch. A schematic of the clutch can be seen in Fig.~\ref{fig:  clutch schematic}.
		
\begin{figure}[!t]
			\subfloat[\label{fig:  configuarion_1}]{%
				\includegraphics[width=0.8in]{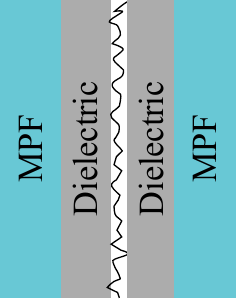}
			}\hfill
			\subfloat[\label{fig:  configuarion_2}]{%
				\includegraphics[width=0.8in]{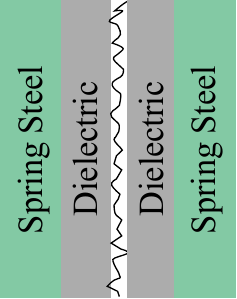}
			}\hfill
			\subfloat[\label{fig:  configuarion_3}]{%
				\includegraphics[width=0.8in]{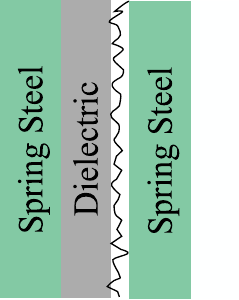}
			}
			\caption{Schematic of the disc configurations. (a) Pair 1. (b) Pair 1. (c) Pair 3.}
			\label{fig:  schematic of discs}	
		\end{figure}

\begin{figure}[!t]
			\centering
			\subfloat[\label{fig:  MPF discs_1}]{%
				\includegraphics[width=1.6in]{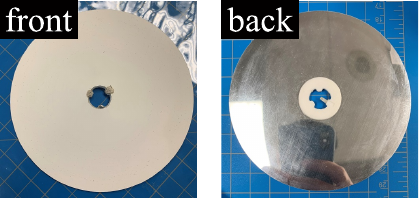}
			}\hspace{0.1in}
			\subfloat[\label{fig:  MPF discs_2}]{%
				\includegraphics[width=1.6in]{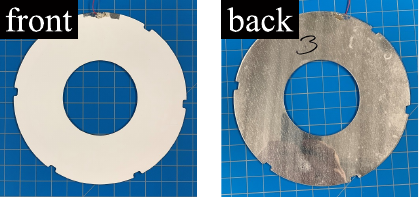}
			}\hfill
			\subfloat[\label{fig:  spring steel discs_1}]{%
				\includegraphics[width=1.6in]{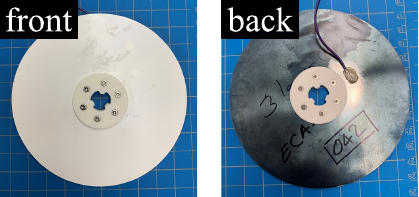}
			}\hspace{0.1in}
			\subfloat[\label{fig:  spring steel discs_2}]{%
				\includegraphics[width=1.6in]{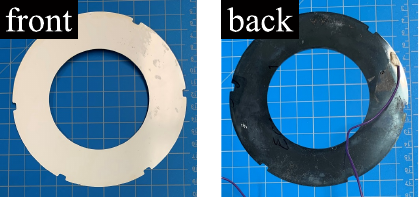}
			}\hfill
			\subfloat[\label{fig:  steel discs_1}]{%
				\includegraphics[width=1.6in]{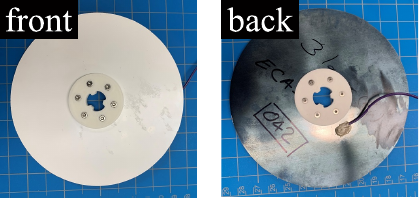}
			}\hspace{0.1in}
			\subfloat[\label{fig:  steel discs_2}]{%
				\includegraphics[width=1.6in]{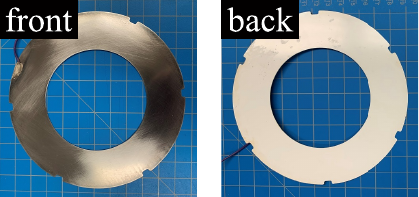}
			}\hfill
			\caption{Configurations of the clutch discs. (a) Pair 1 stator disc. (b) Pair 1 rotor disc. (c) Pair 2 stator disc. (d) Pair 2 rotor disc. (e) Pair 3 stator disc. (f) Pair 3 rotor disc.}	
		\end{figure}

		We considered a number of different configurations of the discs, including different substrate materials and different friction surfaces in our initial feasibility studies and further investigated multiple configurations experientially to arrived at the desired results. The first configuration was based on the method mentioned in \mbox{\cite{diller2018effects}} which had two major issues: high electric resistance and rough sliding friction. The second configuration was a trial to address the high electrical resistance issue of the first configuration. The third configuration was a trial to address the rough friction issue. These three configurations are compared in detail in Section~{\ref{sec: clutch discs selection}}. Fig.~\ref{fig:  schematic of discs} shows a schematic of the disc configurations. The details of the disc configurations are as follow: 
		\begin{enumerate}
			\item A sheet of Metalized Polyester Film (MPF) with 0.127 mm thickness (McMaster Carr, ON, Canada) was used as the substrate. The dielectric was applied to the metalized side of MPF using a 25 $\mu$m wire-wound wet film applicator (GARDCO, FL, USA). The sheet was then cured for 2 hours in $130 ^{\circ}$, rested for 5 hours, and again cured for 2 hours in $130 ^{\circ}$. This procedure was repeated to reach the desired dielectric thickness. After curing the dielectric, the sheet was cut to shape with a laser cutting machine ({Fig.~{\subref*{fig:  MPF discs_1}}} and {Fig.~{\subref*{fig:  MPF discs_2}}} show the discs.). A small area of the dielectric was cleared with Acetone-99.5\% (McMaster Carr, ON, Canada), and a wire was connected to the conductor side of the disc with 8331–Silver Conductive Epoxy (MG Chemicals, BC, Canada).
			
			\item 1095 Spring Steel with a thickness of 0.102 mm (McMaster Carr, ON, Canada) was used as the substrate. The dielectric was applied with the same procedure as described before, and the sheet was sandwiched with thick steel plates and cut to shape with electric discharge machining. A wire was connected to the bare side of the sheets with conductive epoxy. {Fig.~{\subref*{fig:  spring steel discs_1}}} and {Fig.~{\subref*{fig:  spring steel discs_2}}} show the disc pair for the second configuration.
			
			\item A stator disc was made similar to the one built in the second pair but with a thicker dielectric layer. The same rotor disc of configuration \#2 was used for the rotor. However, in this configuration, the steel side of the rotor disc was used for the friction interface. This surface was sanded with 220 grit sandpaper (McMaster Carr, ON, Canada). {Fig.~{\subref*{fig:  steel discs_1}}} and {Fig.~{\subref*{fig:  steel discs_2}}} show the third configuration.  
		\end{enumerate}
	
		The inner diameter of the rotor disc and the outer diameter of the stator disc is 8 cm and 12 cm, respectively. This creates a ring shape friction area with an inner diameter of 8 cm and an outer diameter of 12 cm. Assuming a constant shear stress over the friction ring, the maximum achievable output torque will be,
		\begin{equation}
			T =   \frac{2\pi(r_2^3 - r_1^3) \sigma_{sh}}{3} 
			\label{eq: stress to torque} 
		\end{equation}	
		where $r_1$ and $r_2$ are the inner and outer radii of the friction ring, respectively, and $\sigma_{sh}$ is the shear stress due to electroadhesion.	
			
		\begin{figure}[!b]
			\centering
			\includegraphics[width=.8\linewidth, keepaspectratio, trim=0in 0.3in 0in 0in, clip]{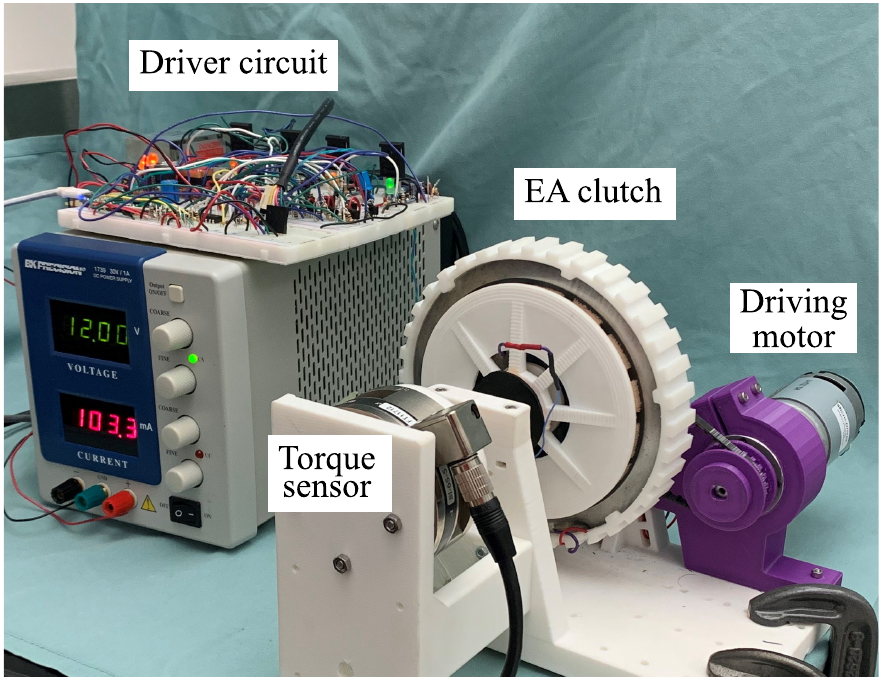}
			\caption{Rotary electroadhesive clutch test-setup}
			\label{fig:  Clutch set up}
		\end{figure}

	\subsection{Driver circuit design}
		In order to activate the clutch, a high-voltage potential is needed across the clutch discs. In our setup, a high voltage PHV 12 DC-DC transformer (Bellnix Co., Saitama, Japan) was utilized to feed the circuit with the required voltage. An R-2R digital-to-analog converter was used to adjust the reference voltage of the transformer. In order to change the polarity of the activation of the clutch, a high-voltage IGBT (Infineon Technologies, Neubiberg, Germany) H-bridge was used. This allowed activation of the clutch under an Alternating Current (AC) waveform as well as Direct Current (DC). For safety reasons, the high-voltage part of the driver circuit was isolated from the logic part using optocouplers and isolated DC-DC converters. An Arduino microcontroller board was utilized to interface the driver with a computer. 
			
	\subsection{Experimental setup}
		An experimental setup was fabricated to validate the performance of each pair of discs. Except for the clutch discs, the shaft, and the standard parts, other parts of the setup and EA clutch were fabricated with 3D printing. The diameter of the fabricated EA clutch used in the setup was 15 cm, the thickness of a pair of clutch discs was 0.36 mm, the thickness of the complete clutch unit, including the pulley and slipring, was 73 mm, and the length of the output shaft of the clutch was 166 mm. The mass of a pair of  clutch discs for the configuration \#2 and \#3 was 20 g and for the configuration \#1 was less. The mass of the complete EA clutch (excluding the base frame, motor, and torque sensor) was around 420 g. It should be noted that the housing and other main components of the proposed EA clutch were not designed to be weight-efficient. Since the setup was developed for validation purposes, it was slightly over-designed with larger space between the components to allow future expansion and modification of the setup. The point to highlight is the lightness of the core components of the clutch i.e., the clutch disc. This allows to integrate a number of clutch discs in the same cover for a much higher torque-to-weight ratio of the EA clutch. The overall weight of the EA clutch can be further reduced by the optimal design of the components.
		
		The clutch was driven using a DC gear motor (Micro-Drives M4870U with 1:294 gear ratio) controlled by a Maxon EPOS2 under speed control mode with a constant speed. The output of the clutch was connected to a torque sensor (ATI Industrial Automation, NC, USA). The stator disc was directly connected to the driver circuit, and the rotor disc was connected to the driver circuit through a slip-ring to allow free rotation of the disc. Fig.~\ref{fig:  Clutch set up} shows the fabricated experimental setup.
		
		Data acquisition and control of the clutch were made using QuaRC real-time control software (Quanser, ON, Canada) in the MATLAB Simulink environment because of versatility and easy-to-use connection establishment with the experimental setup. The frequency of data acquisition was set to 500 Hz for validating the clutch performance under AC activation signals and for validating pHRI.

\section{Clutch Discs Selection}
\label{sec: clutch discs selection}
\begin{figure}[!t]
		\centering
		\includegraphics[width=\linewidth, keepaspectratio]{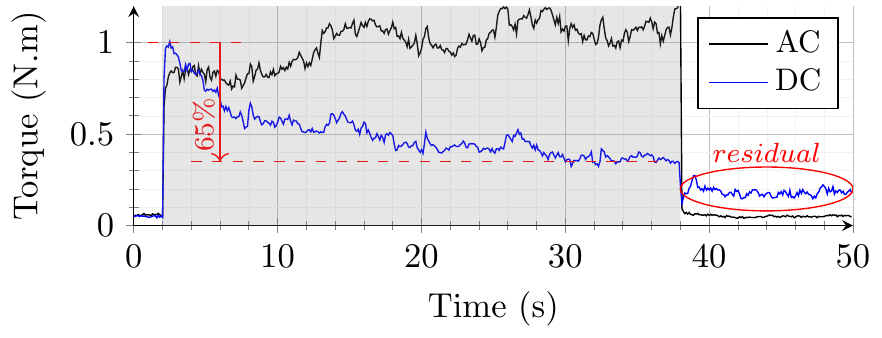}
		\caption{Output torque for dielectric against dielectric friction disc pair \#1 (blue graph) and \#2 (black graph) under DC and AC (50 Hz) activations. The active region is shaded in gray.}
		\label{fig: dielectric on dielectric}
	\end{figure}

	The performance of the clutch was evaluated with each configuration explained above. Configuration \#1 was tested in the test-setup with a constant 400 volts DC activation. The blue graph in Fig.~\ref{fig: dielectric on dielectric} shows the results for configuration \#1. As seen in the figure, the torque degrades up to almost 65\% with the passage of time. In addition, the residual torque can be seen after the deactivation of the clutch. The torque degradation and residual torque result from the charge build-up in the dielectric due to the constant electric field \cite{guo2017experimental}. In this work, we addressed the charge build-up issue by activating the clutch with an AC waveform \cite{chen2020time}. Activating an EA clutch with an AC waveform required a different disc configuration since the conductor film of MPF was too thin, resulting in significant resistance. This resistance prevented the required current flow due to charge-discharge cycles needed under AC activation. The higher resistance and the heat build-up can damage the conductor area close to the wire-to-disc connection on MPF.
	
	In order to solve the above-mentioned problem, configuration \#2, which was made of a thicker layer of conductor, was fabricated. Configuration \#2 was evaluated using an AC square waveform activation with an amplitude of 400 V and frequency of 50 Hz. The black graph in Fig.~\ref{fig: dielectric on dielectric} shows the output torque of configuration \#2 under AC activation. It can be seen that the issue of torque degradation and residual torque after deactivation are substantially improved. However, the clutch torque fluctuates significantly (Torque = 1$\pm$0.11 N.m). This phenomenon is because of the rough friction that results from the same materials sliding against each other (dielectric against dielectric). Atomic and molecular bonds may form between the surfaces of the dielectric, which results in a high coefficient of friction affecting the smoothness of the torque transfer and, therefore, higher shear force. This can be good for locking mechanisms but not for a tunable system such as the one proposed here. 

\begin{figure}[!t]
		\centering
		\includegraphics[width=\linewidth, keepaspectratio]{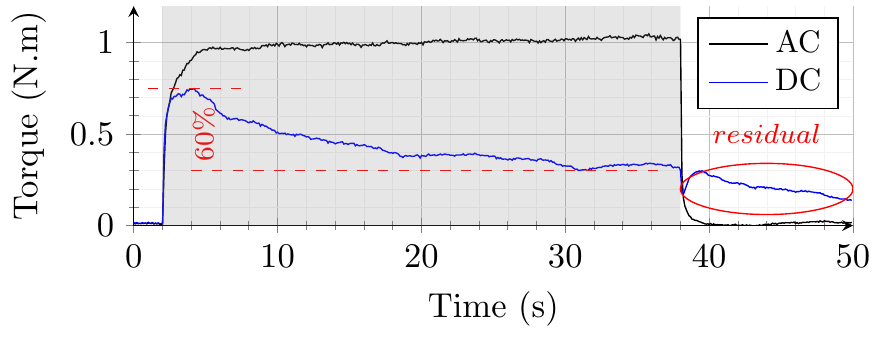}
		\caption{Output torque for disc pair \#3 (dielectric against steel friction) under DC and AC activations. The active region is shaded in gray.}
		\label{fig: dielectric on steel}
	\end{figure}

	In order to address torque fluctuations, configuration \#3 was proposed, in which dielectric were placed against steel (not another layer of dielectric). Fig.~\ref{fig: dielectric on steel} shows the results for both AC and DC activations in configuration \#3. As seen, the torque fluctuation is almost 6 times less than that in configuration \#2 (Torque = 1$\pm$0.02 N.m for configuration \#3 vs. Torque = 1$\pm$0.11 N.m for configuration \#2). It is clear that configuration \#3 with an AC activation offers the best performance with no obvious issue. Configuration \#3 was selected for further investigation and further development of the experimental setup.

\section{Performance Evaluation and Modeling}
\label{sec: performance  eval and modeling}
	In this section, the proposed configuration \#3 is evaluated through a series of tests using both AC and DC activations. 
	
	\subsection{Experiment condition}
		In these tests, the rotor of the clutch was rotated with a constant velocity of 5 RPM in each test and the clutch was activated for a period. The maximum torque of the motor was set to be more than the maximum clutch torque to allow clutch discs to continuously slide against each other when the clutch was active. After each DC activation test, the clutch was activated with 260 volts AC waveform for three minutes to depolarize the dielectric before the next test. All experiments were carried out at room temperature (20$^{\circ}$C-25$^{\circ}$C) and humidity (30\%-50\%). The procedure was repeated for different activation voltages, and the output torque of the clutch was recorded.
	
	\subsection{AC activation model}		
\begin{figure}[t]
			\centering
			\includegraphics[width=\linewidth, keepaspectratio]{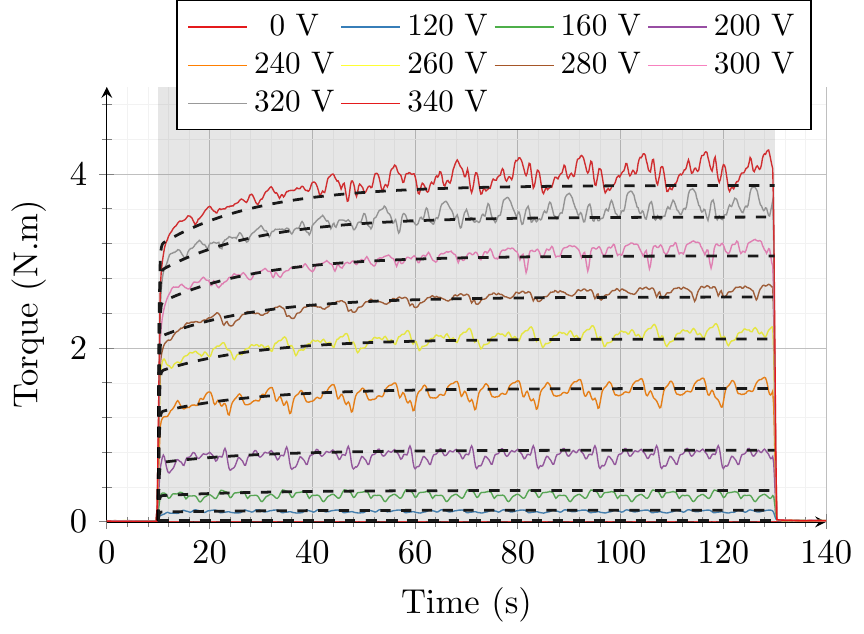}
			\caption{Clutch performance under 300 Hz AC activation. The clutch was activated at t = 10 s and was deactivated at t = 130 s. The active region is shaded in gray. Dashed lines are the torque predicted by the model.}
			\label{fig: res_AC activation}
		\end{figure}

		{Fig.~\ref{fig: res_AC activation}} shows the results for 300 Hz AC activation. The amplitude of the square activation signal was in the range of 0~V to 340~V. The results support the desired functionality for the proposed design as the torque remains almost constant during the activation period without degradation. The fast transition dynamics for the major part of the torque, along with slow transition dynamics for the minor part of the torque, can be seen at the beginning of the diagram. 
		
\begin{figure}[t]
			\centering 
			\includegraphics[width=\linewidth, keepaspectratio]{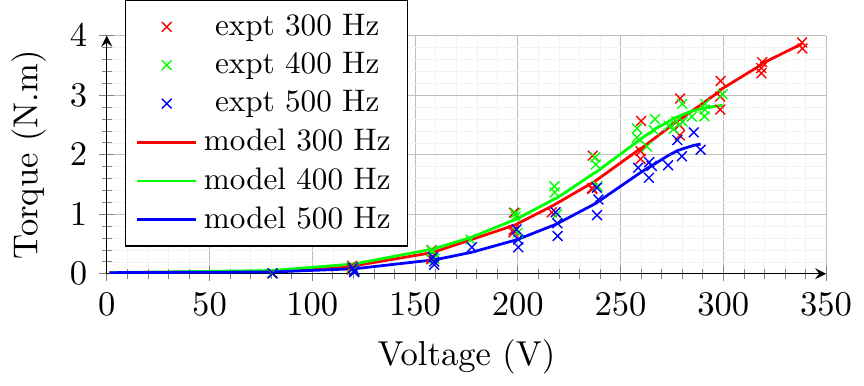}
			\caption{Averaged output torque (points) and proposed models (solid lines) for different activation voltages in AC mode.}
			\label{fig: res_AC voltage effect}
		\end{figure}

		As observed in {Fig.~\ref{fig: res_AC activation}}, the transmission torque of the EA clutch increases with the increase of the voltage. This can be seen more clearly in {Fig.~\ref{fig: res_AC voltage effect}} which shows the average torque versus the activation voltage for the AC activation signal for different frequencies of 300 Hz, 400 Hz, and 500 Hz. To study potential uncertainties and assess the effect of model, we recorded the results on three different days to account for any changes in the friction properties due to temperature and humidity changes. As seen in {Fig.~\ref{fig: res_AC voltage effect}}, some saturation can be seen for activation voltages above 250~V roughly, which was expected based on the literature on electroadhesion saturation \mbox{\cite{guo2017experimental,koh2014experimental}}. The activation voltage for which torque saturation occurs reduces with the increase of the frequency of the activation signal. The occurrence of torque saturation at lower activation voltage prevents achieving higher torque when the frequency is high. The torque saturation is caused by the dielectric leak, which increases exponentially in high voltages, and possibly because of the accumulation of residual charges, which prevent higher electric charge on the discs. As the frequency of the signal increases, the resistance of the dielectric and the reactance of the clutch (which acts similar to a capacitor) reduces, resulting in a higher current leak through the dielectric \mbox{\cite{guo2019electroadhesion}}. The lower is the frequency, the higher is the maximum torque that leads to dielectric breakdown. 
		
		\begin{table}[b]
			\caption{Identified parameters of the models
			\label{tab: model params}}
			\centering
			\begin{tabular}{ c c c c c c}
				\hline
				\hline
				freq (Hz) & error (N.m) & K 			 & n 	& c (V) 	 & $\delta$\\  \hline 
				300 & $\num{1.79e-1}$ & $\num{1.01e-4}$  & 3.89 & 306.51 &  38.89 \\
				400 & $\num{1.40e-1}$ & $\num{2.45e-4}$  & 3.74 & 279.49 &  25.83 \\ 
				500 & $\num{1.27e-1}$ & $\num{7.3e-5}$   & 3.88 & 281.92 &  6.67 \\ 			
			\end{tabular}
		\end{table}
		
		On the other hand, higher frequency provides a smoother output torque. Therefore, there is a trade-off between the maximum achievable torque and the smoothness of the torque. As seen in the same figure, the maximum torque achieved using one pair of clutch discs and an AC activation signal with a frequency of 300 Hz is 3.9 N.m. This amount of torque is sufficient for most upper-limb human robot interaction applications as will be shown in Section {\ref{sec: HRI experiment}}. We also noted that the frequency of the activation signal affected the sound noise generated by the clutch.
		
		Using the general model of shear stress presented in \eqref{eq: EA shear stress}, the following model is proposed to explain and predict the behavior of the designed EA clutch.
		\begin{equation}
			\overline{\sigma^{AC}_{sh}}(V) = C_f\Big(\sigma_0 +  	\frac{K\epsilon_r\epsilon_0 	V_{eff}^{n_1}(V)}{d^2}\Big) 
			\label{eq: EA shear stress AC average}
		\end{equation}
		where $C_f$ is the friction coefficient, $\sigma_0$ = 0.02 N.m\textsuperscript{-2} is the permanent pressure due to pressure springs, $K$ is a correction factor, $\epsilon_r = 35$ is the relative permittivity of the dielectric, $\epsilon_0$ is the permittivity of vacuum, $n_1$ is a coefficient defining the intensity of the effect of voltage on the shear stress, $d = 80 \; \mu$m is the dielectric thickness, and $V_{eff}$ is the effective voltage that is calculated as follows,
		\begin{equation}
			V_{eff}(V) =  V-\ln{\Big(1+e^{\delta(V-c)}\Big)}  	
			\label{eq: effective voltage}	
		\end{equation}		
		in that $V$, $c$, $\delta$ are input voltage, saturation voltage, and sharpness factor of the saturation function, respectively. 
		
		The hyperparameters of the model in {\eqref{eq: EA shear stress AC average}} and {\eqref{eq: effective voltage}} depend on the frequency of the activation signal and the fabrication process. Therefore, these parameters should be calculated through system identification using the torque-voltage experimental results (see {Fig.~\ref{fig: res_AC voltage effect}}). To this effect, individual models were fitted to the data for each activation frequency using the least-squares method. The torques of the EA clutch predicted using the above-mentioned models as a function of the activation voltage are shown in {Fig.~\ref{fig: res_AC voltage effect}} using solid lines. The values of the hyperparameters and estimation errors of the models are shown in \mbox{{Table~\ref{tab: model params}}}.	
	
		Also, as seen, the predicted torque by the model is related to the voltage by a exponent of 3.8, which does not comply with the theoretical model presented in {\eqref{eq: EA shear stress}} that is related to the square of the voltage. This is due to the fact that the actual effect of voltage on the transmission torque is more than what can be captured using \eqref{eq: EA shear stress}. This observation is somewhat intuitive because in \eqref{eq: EA shear stress}, only the effect of the shear stress due to the attraction force between the conductors is included. However, in practice, the attraction force between the clutch discs is due to the attraction of the conductors as well as the attraction between the dielectric particles on the stator disc and the rotor disc (as mentioned in \eqref{eq: EA shear stress isolator}), which are both related to the activation voltage. The latter attraction force is difficult to be modeled without the molecular knowledge of the dielectric. In addition, the relation of the friction force to the pressure may also be nonlinear, leading to a nonlinear dependency of the torque to the applied voltage.
		
		Another important observation based on the results in {Fig.~\ref{fig: res_AC activation}} is the finite rise time for the torque. This time-dependent behavior is not due to the sluggish dielectric polarization because the polarization part with slow dynamics was eliminated under a high-frequency AC activation regime \cite{chen2020time}. Therefore, the time-dependent behavior is believed to be due to mechanical characteristics of the system, including dielectric surface texture and friction dynamics. Because these effects cannot be investigated without the knowledge of dielectric surface texture, two independent exponential activation functions with independent time constants were added to the time-independent torque model {\eqref{eq: effective voltage}} to model the torque dynamics that immediately appear after the activation voltage is applied, as shown below,
		\begin{equation}
			\sigma_{sh}^{AC}(t,V) = 		\overline{\sigma_{sh}^{AC}}(V)\Big(\alpha\big(1-e^{-t/\tau_1}\big) + 	(1-\alpha)\big(1-e^{-t/\tau_2}\big)\Big)
			\label{eq: EA shear stress AC time dependent} 
		\end{equation}
		where $\tau_1$ and $\tau_2$ are time constants of the faster and slower activation functions, respectively, and  $\alpha$ is the weighting factor that defines the portion of the output torque that reacts with time constant $\tau_1$. These values were determined to be 0.14 s, 18.70 s, and 0.82, respectively, through system identification using {\eqref{eq: EA shear stress AC time dependent}} to the experimental torque-time data. According to obtained dynamic model, 82\% of the torque reacts to the input signal with a time constant of 0.14 s, and the remaining 18\% of the torque reacts with a slower dynamic with a time constant of 18.7 s. The results from this model for various activation voltages are shown in Fig.~\ref{fig: res_AC activation} using dashed lines. It can be seen that the proposed dynamic model has a high agreement with the experimental recordings. The results also verify that the two activation functions with short and large time constants affect the dynamics of the system.
		
	\subsection{Frequency effect}
\begin{figure}[t]
			\centering
			\includegraphics[width=\linewidth, keepaspectratio]{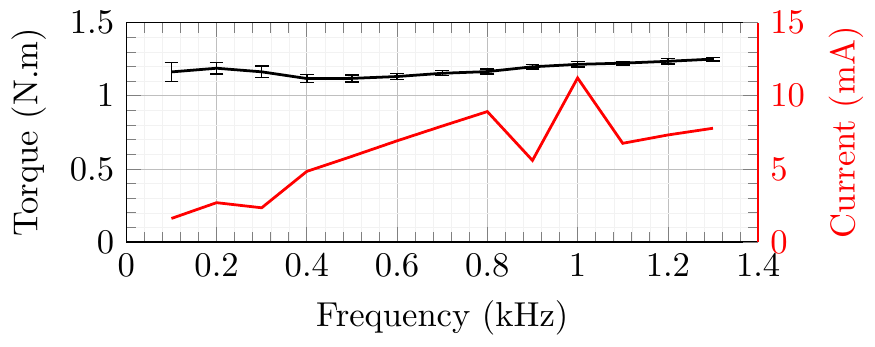}
			\caption{Output torque and input current for different frequencies of the activation voltage with 180~V amplitude.}
			\label{fig: res freq effect}
		\end{figure}

		The output torque of the EA clutch and associated input current were obtained for an activation voltage with various frequencies and a constant amplitude  (180 AC volts). {Fig.~\ref{fig: res freq effect}} shows the effect of the activation frequency on the output torque and the input current. The output torque diagram is the average of four repeated tests, and the error bars are standard deviations. Results are for frequencies ranging from 100 Hz to 1300 Hz. It can be seen that the output torque reduces with the increase of the frequency so long as the frequency remains below 500 Hz. However, when the frequency is above 500 Hz, the torque increases with a small slope with the frequency increase. There are uncertainties (mostly due to friction and the changes in the surface finish with the age of the clutch) that have effects on the output torque, as can be seen in {Fig.~\ref{fig: res_AC voltage effect}}. However, there is no solid explanation for the torque increase with the frequency increase when the frequency is above 500 Hz. The most significant effect of the activation frequency is on the saturation voltage, which drops as the frequency increases (see {Fig.~\ref{fig: res_AC voltage effect}}). Also, it can be seen that the standard deviations are larger at lower frequencies, which shows higher torque fluctuations at lower frequencies. The fluctuations remain almost constant over 500 Hz. The lower fluctuations and the lower sound noise were among the reasons that a frequency of 500 Hz was selected for the subsequent tests.
		
		In addition, it can be seen that although the torque remains constant with the increase of the frequency, the required current increases in a non-monotonic manner. The increase of the current with the increase of the frequency can be explained as follows: First, in each period of the activation signal, the EA clutch would charge, discharge, and charge again with opposite polarity, and discharge again. Assuming a small unit of energy is exchanged in each cycle, with the increase of the frequency, the number of charge-discharge cycles increases; resulting in increased heat dissipation in a unit of time \cite{hinchet2020high}. Second, the resistance of the dielectric reduces with the increase of frequency which leads to a higher charge leak through the dielectric. The same relation between frequency and power consumption is also reported in \cite{chandrakasan1992low} for CMOS integrated circuits.
		
	\subsection{DC activation model}
		Although in this paper we proposed the use of AC voltage to address the torque performance of the EA clutch, it is necessary to study and understand the behaviour of the EA clutch under DC activations as well. 
		
\begin{figure}[b]
			\includegraphics[width=\linewidth, keepaspectratio]{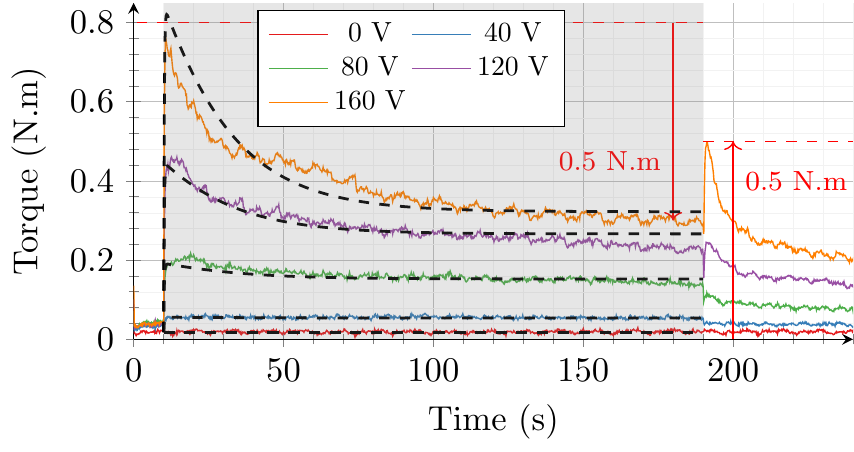}
			\caption{Clutch performance under a DC activation. The clutch was activated at t = 10 s and was deactivated at t = 190 s. The active region is shaded in gray. Dashed lines are the torque predicted by the model.}
			\label{fig: res_DC activation}
		\end{figure}

		To this end, Fig.~\ref{fig: res_DC activation} shows the results for different DC activation voltages. As expected, the EA clutch torque increases with the increase of the activation voltage, which is because of the increase of the electric field with voltage. In addition, it can be seen that the maximum torque of the EA clutch is reached at the beginning of the activation cycle. After that, the torque starts to degrade exponentially until it reaches a saturated value \cite{chen2020time}.
		
		As explained in Section \ref{sec: Concepts}, the attraction force between the clutch discs results from the charge accumulated on the conductors as well as the polarization of the dielectric that is placed in an electric field. The dielectric polarization in materials, such as barium titanate, is mostly because of the orientational polarization (which has a short time constant of less than 10 $\mu$s) and space charge polarization (which has a larger time constant up to hundreds of seconds) \cite{guo2016optimization}. Although, in orientational polarization, only the direction of the dipoles, which does not affect the primary electric field, changes, in space charge polarization, electrons migrate to the surface of the dielectric and may get stuck on the surface. This results in charge build-up on the surface of the dielectric and develops an electric field that counteracts the primary electric field, leading to a resultant electric field that is formulated as follows \cite{guo2016optimization},
		\begin{equation}
			\va{E}(t) = \va{E_p} + \va{E_r}(t)    
			\label{eq: Electric field1}
		\end{equation}
		where $\va{E_p}$ is the primary electric field due to the applied voltage and $\va{E_r}(t)$ is the electric field due to charge build-up inside the dielectric because of the dielectric polarization.
		
		In addition to the torque degradation issue, there exists residual torque after the deactivation of the clutch. In Fig.~\ref{fig: res_DC activation}, the residual torque can be seen after t = 190 s. With the removal of the primary electric field, the resulting electric field, which used to be the summation of $\va{E_p}$ and $\va{E_r}(t)$ will change to only $\va{E_r}(t)$. This activates the EA clutch with a reverse polarity which decays with the depolarization time constant of the dielectric. The magnitude of the resultant electric field can be formulated as follows,
		\begin{subnumcases}{\label{eq: electric field2} |E(t)| = }
			|E_p| - |E_r(t)| & activated \\
			|E_r(t)| 	& deactivated
		\end{subnumcases}
	
		Following the removal of the primary electric field, the electric field created by the residual charges keeps the clutch engaged. The initial value of the torque after the deactivation of the clutch is equal to the amount of degradation of the torque at the end of the active cycle (see Fig.\ref{fig: res_DC activation}), verifying the dynamics explained in \eqref{eq: electric field2}.
		
		In order to model the dynamics of the clutch in DC activation, an exponential degradation function is multiplied by the AC model given in \eqref{eq: EA shear stress AC time dependent}  as shown below,
		\begin{equation}
			\label{eq: EA shear stress DC}
			\sigma_{sh}^{DC}(t,V) = \sigma_{sh}^{AC}(t,V) 	\Big(1-\big(1-e^{-t/\tau_d}\big)C_s 	V^{n_2}\Big)   
		\end{equation}
		where $\tau_d$ is the time constant of the degradation function. The term $Cs V^{n_2}$ is added to adjust the effect of the voltage on the saturation of the settled torque. The unknown parameters are calculated by fitting the proposed model to the experimental data using the least-squares method. As such, $\tau_d$, $n_2$, $C_s$, were determined to be 20.9 s, 1.17, and 0.0017, respectively, which says that the torque degrades by 63 \% in the first 20 seconds of activation. Therefore, it can be said that the output torque of the EA clutch is subjected to significant degradation under DC activation. The results from the proposed model are shown by the dashed lines in Fig.~\ref{fig: res_DC activation}.

\begin{figure}[!b]
			\centering
			\includegraphics[width=.7\linewidth, keepaspectratio]{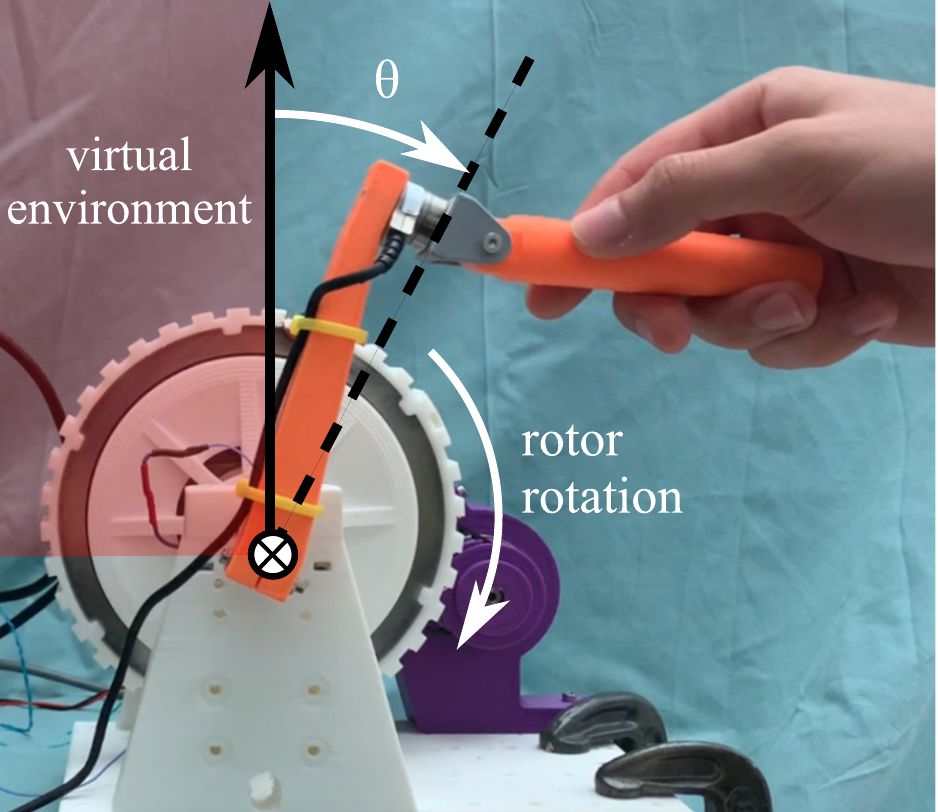}
			\caption{One-DoF haptic device for pHRI}
			\label{fig: HRIexpt setup}
		\end{figure}

%
%
%
		The experimental setup shown in {Fig.~\ref{fig:  Clutch set up}} was modified into a one-DoF haptic device, as shown in {Fig.~\ref{fig: HRIexpt setup}}. An 11 cm arm was connected to the output shaft of the clutch. A handle was connected to the end of the arm with a revolute joint. An ATI nano 17 force/torque sensor was mounted between the handle and the arm to measure the interaction forces. An incremental encoder (CUI AMT10 with 2048 counts per revolution) was used to measure the joint angle of the haptic arm.
		
\begin{figure}[!b]
		 	\centering
		 	\includegraphics[width=0.9\linewidth, keepaspectratio]{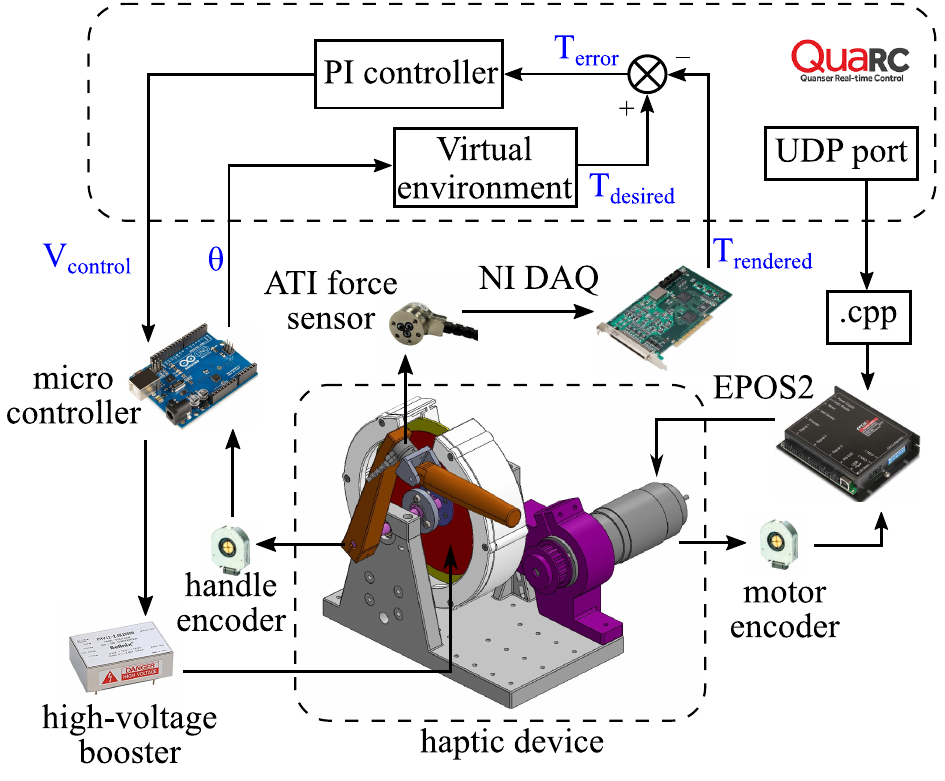}
		 	\caption{Schematic of the haptic device and the controller architecture.}
		 	\label{fig: control diagram}
		\end{figure}

		A simple impedance control based on a proportional-integral controller was implemented to render a virtual environment using a virtual rotational spring around the output shaft of the clutch with the stiffness of K. The virtual environment covered the space where $\theta \in [-\pi/2, 0]$. The virtual environment applies different values of resistive torque depending on its stiffness and the angle of the haptic arm. For data acquisition and control, we used QuaRC real-time control software (Quanser, ON, Canada) in the MATLAB Simulink environment with a sampling frequency of 500 Hz. {Fig~\ref{fig: control diagram}} shows the schematic of the system.
		 
		Three different stiffness values to resemble a soft (K = 1 N.m/rad), medium (K = 2 N.m/rad), and hard (K = 4 N.m/rad) environment were implemented in the experiments. The user interacted with the environment by moving the haptic handle in a reciprocating fashion during the experiments. In the first experiment, the controller parameters were manually tuned to $K_P = 200$ and $K_I = 10000$, and the rotor of the clutch was rotated with a constant velocity of 15 RPM so that the haptic device could apply resistive and assistive forces in the positive direction. In the second experiment, the input velocity of the clutch was set to zero, and the controller parameters were manually tuned to $K_P = 800$ and $K_I = 0$. Because of the zero input velocity, the haptic device could only apply a resistive force. Thus, an automatic switch was implemented in the code to release the clutch when the handle was getting out of the environment to avoid resistive force on exit.
		
		\begin{figure}[!b]
			\raggedleft
			\subfloat[]{\label{fig: X position_11}
				\includegraphics[keepaspectratio, right]{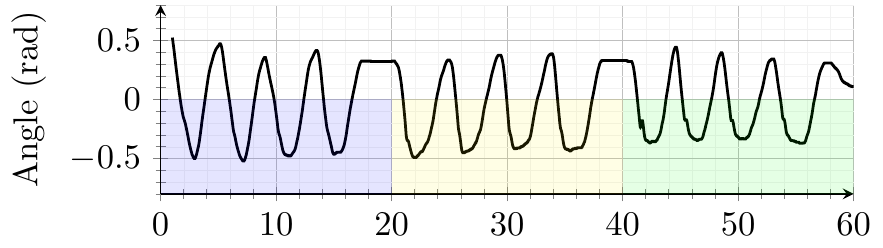}
			} \hfil
			\raggedleft
			\subfloat[]{\label{fig: X force_11}
				\includegraphics[keepaspectratio, right]{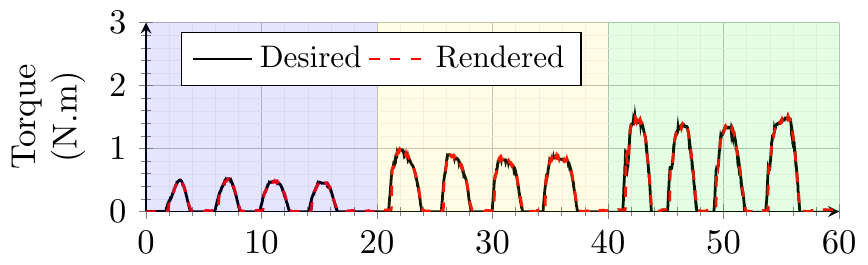}
			} \hfil
			\subfloat[]{\label{fig: stiffness_11}
				\includegraphics[keepaspectratio, right]{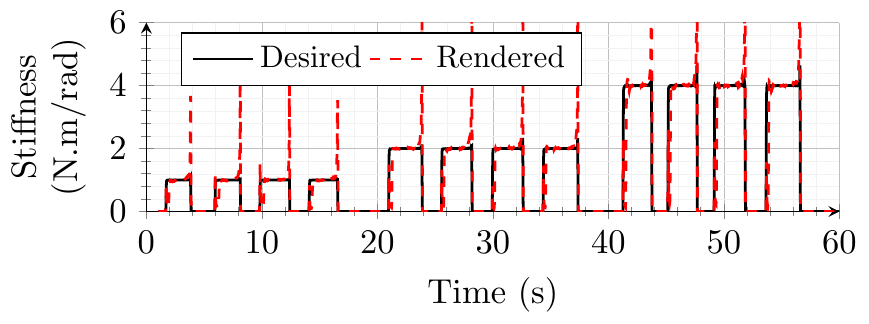}
			} \hfil
			\caption{Physical human-robot interaction experimental results for the clutch input velocity of 15~RPM. (a) Handle angle. (b) Desired and rendered torque. (c) Desired and rendered stiffness.}
			\label{fig: HRIexpt_11}
		\end{figure}
		
	\subsection{Results and discussion}	
		Results of the first pHRI experiment can be seen in {Fig.~\ref{fig: HRIexpt_11}}. Fig.~{\subref*{fig: X position_11}} shows the angle of the haptic handle. As seen, the rendered torque of the haptic device perfectly tracks the desired environmental torque in both forward and backward motions. The clutch is able to track the desired torque in backward (positive) direction as long as the velocity of the operator's hand is lower than the rotation velocity of the clutch rotor~\mbox{\cite{pisetskiy2021high}}. The same performance can be seen in Fig.~{\subref*{fig: stiffness_11}}, where the rendered stiffness tracks the desired stiffness. Spikes in the rendered stiffness are due to the short time delay (in the order of 50 milliseconds) of the control signal which is related to the time lag of the high-voltage booster and the time lag due to the integral part of the controller used to activate the clutch.
			
		\begin{figure}[!b]
			\raggedleft
			\subfloat[]{\label{fig: X position_12}
				\includegraphics[keepaspectratio,right]{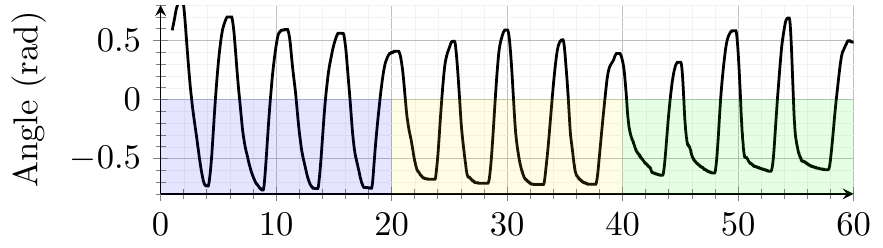}
			} \hfil
			\subfloat[]{\label{fig: X force_12}
				\includegraphics[keepaspectratio,right]{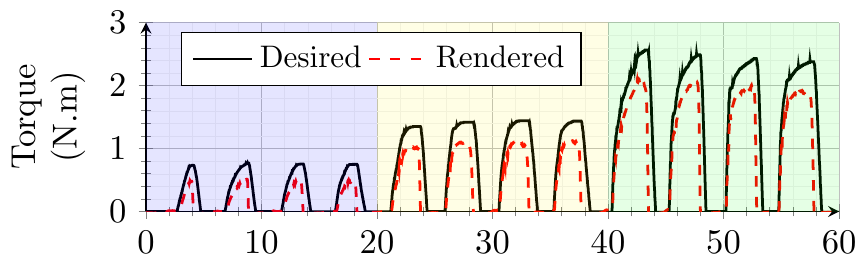}
			} \hfil
			\subfloat[]{\label{fig: stiffness_12}
				\includegraphics[keepaspectratio,right]{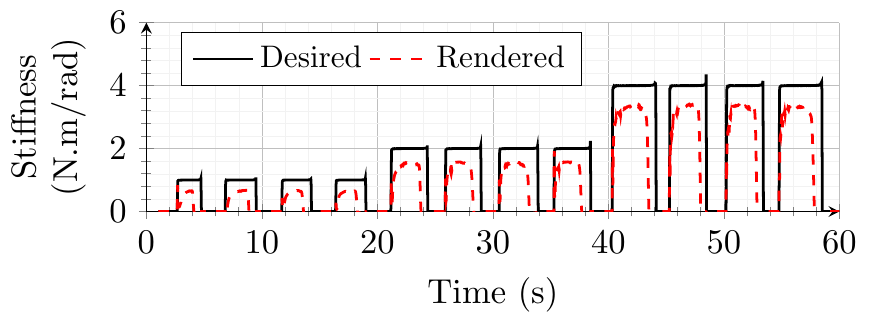}
			} \hfil
			\caption{Physical human-robot interaction experimental results for the clutch input velocity of 0. (a) Handle angle. (b) Desired and rendered torque. (c) Desired and rendered stiffness.}
			\label{fig: HRIexpt_12}
		\end{figure}	
		
		The second set of experimental results are shown in {Fig.~\ref{fig: HRIexpt_12}}. Fig.~{\subref*{fig: X force_12}} shows the desired and rendered torque due to the stiffness of the virtual environment. As ward seen, the rendered resistive torque tracks the desired trajectory in forward (negative) direction of the handle, which pushes the handle into the environment. A slight deviation from the desired torque is due to the simplicity of the proportional controller used in this experiment. The integral term of the controller was eliminated to avoid the accumulation of the control action due to steady-state errors. For the positive velocity of the handle, the rendered force drops to zero because of the automatic release switch integrated into the code. The same behaviour can be seen in the rendered stiffness shown in Fig.~{\subref*{fig: stiffness_12}}.
	
		The proposed rotary EA clutch offers a much higher torque-to-weight ratio and torque-to-power ratio than common adjustable clutches. Unlike MR clutches that need a heavy coil, permanent magnet, and MR fluid to operate \mbox{\cite{moghani2016design}}, an EA clutch only needs a pair of clutch discs covered with a thin layer of dielectric. Our study shows that a pair of discs that weigh around 20 grams can transfer up to 3.9 N.m of torque. The results are important as they inform the feasibility of increasing the number of clutch discs in parallel \mbox{\cite{ramachandran2019all}} without a significant increase in the total weight of the clutch. Up to 20 N.m of torque can be achieved for a clutch with around 500 grams. This range of torque is suitable for direct implementation in mobile systems interacting with humans such as robotic walkers \mbox{\cite{hirata2006motion}}.
		
		In terms of the design complexity, an EA clutch is much simpler to design than ER and MR clutches since an EA clutch does not require sealing to keep the operational fluid inside the clutch \mbox{\cite{davidson2018electrorheological, moghani2016design}}. However, as mentioned in Section {\ref{sec: clutch design}}, the process of applying the dielectric on the clutch discs is time-consuming due to the need for multiple repetitions and high fabrication accuracy. Moreover, the output torque of the EA clutch is more prone to variations due to the variations of the dry friction compared to ER and MR clutches that use viscous friction.

	\subsection{Limitations and future works}
		Although the proposed clutch showed great potential for pHRI applications, the transparency of the interaction can be improved using a robust or adaptive controller that consider the uncertainties of the system. Another shortcoming of the proposed EA clutch is that it generates a sound noise with AC activation. The noise necessitates the need for further vibration analysis of the system and possible inclusion of acoustic dampers in the future design. While major wear and heat generation were not noticed in our study, these two factors have to be investigated for long-term use of EA clutches. Lastly, the effect of operative conditions such as temperature and humidity should be investigated for extensive use of this technology.
		
\section{Conclusions}
\label{sec: conclusions}
	A novel rotary electroadhesive clutch was designed and fabricated for human-robot applications. The proposed method enabled torque controllability by adjusting the activating voltage. Three different materials were investigated for the clutch discs, and the performance of the clutch was validated through extensive experiments including physical human-robot interaction. Two main issues of torque degradation with time and residual torque after deactivation were addressed by activating the clutch with alternating current. In addition, a new computational model was proposed to estimate the dynamics of the clutch torque for both direct and alternating current activations. Such models are essential for dynamic closed-loop control. The proposed clutch demonstrated a torque-power consumption ratio of six times more than a commercial magnetic particle clutch. The proposed clutch showed great potential for application in safe passive actuators such as those required for lightweight and low-power consumption robotic systems in the future.

\ifCLASSOPTIONcaptionsoff
  \newpage
\fi


\bibliographystyle{IEEEtran}
\bibliography{biblography_TMECH}

\begin{IEEEbiography}
	[{\includegraphics[width=1in,height=1.25in,clip,keepaspectratio]{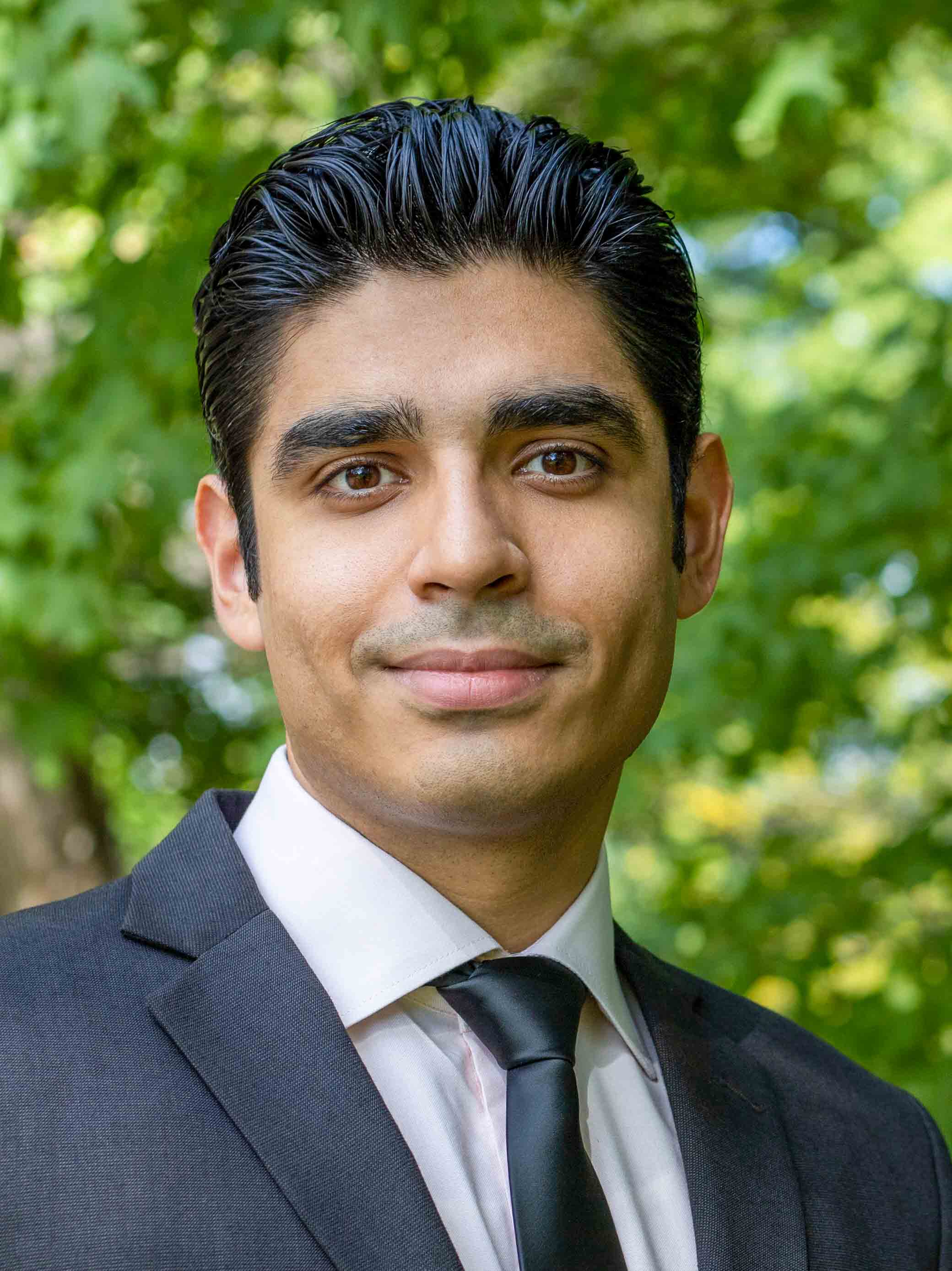}}]{Navid Feizi} (Graduate Student Member, IEEE) received the B.Sc. degree from Shiraz University, Shiraz, Iran, in 2016, and the M.Sc. degree from the Sharif University of Technology, Tehran, Iran, in 2019, both in mechanical engineering. He is currently working toward the Ph.D. degree with the School of Biomedical Engineering (Mechatronics), Western University,	London, ON, Canada.
	
	He is currently a Research Assistant at Canadian Surgical Technologies and Advanced Robotics (CSTAR), Lawson Health Research Institute, London, ON, Canada. He is also a Trainee in effective systems for procedure specific healthcare simulation at CSTAR. His research interests include human–robot interactions, rehabilitation and surgical	robotics, haptics, teleoperation, and smart actuators.
\end{IEEEbiography}

\begin{IEEEbiography}
	[{\includegraphics[width=1in,height=1.25in,clip,keepaspectratio]{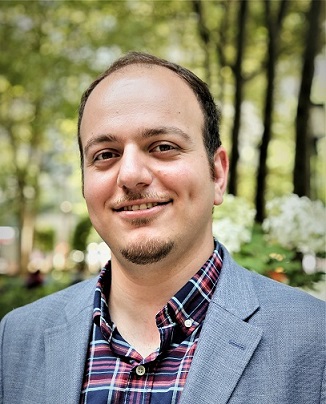}}]{S. Farokh Atashzar} (Senior Member, IEEE) has received a Ph.D. degree in Electrical and Computer Engineering from the University of Western Ontario, Canada, in 2017. He is currently an Assistant Professor with New York University (NYU), New York, NY, USA, jointly appointed with the Department of Electrical and Computer Engineering and Mechanical and Aerospace Engineering. He is also affiliated with NYU WIRELESS, New York, NY, USA, and NYU Center for Urban Science and Progress, New York, NY, USA, and leads Medical Robotics and Interactive Intelligent Technologies (MERIIT) Lab, NYU, and the activities of the lab are funded by the US National Science Foundation. Prior to joining NYU, he was a Postdoctoral Scientist with Imperial College London, U.K. His research interests include a human-machine interface, haptics, human-centered robotics, biosignal processing, deep learning, and nonlinear control. Prof. Atashzar was the recipient of several awards, including the 2021 Outstanding Associate Editor of IEEE ROBOTICS AND AUTOMATION LETTERS. He is an Associate Editor for IEEE TRANSACTIONS ON ROBOTICS and IEEE Robotics and Automation Letters. 
\end{IEEEbiography}

\begin{IEEEbiography}
	[{\includegraphics[width=1in,height=1.25in,clip,keepaspectratio]{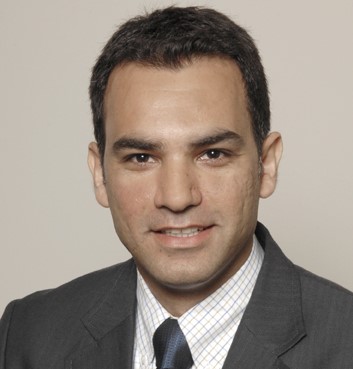}}]{Mehrdad R. Kermani} (Member, IEEE) received his PhD in Electrical and Computer Engineering from The University of Western Ontario, London, ON, Canada in 2005. He is currently a Professor with the Department of Electrical and Computer Engineering and the Director of Advanced Robotics and Mechatronic Systems laboratory, at Western University. His expertise and research interests include human–robot collaborations, robotic grasping, compliant actuations, agricultural robotics, and the use of smart materials for developing human-centric technologies in robotics. Dr. Kermani has served as the Co-Chair of a number IEEE conferences, including ICRA and the selection panel of multiple funding agencies. He is currently an Associate Editor for the IEEE Robotic and Automation Letters.
\end{IEEEbiography}

\begin{IEEEbiography}
	[{\includegraphics[width=1in,height=1.25in,clip,keepaspectratio]{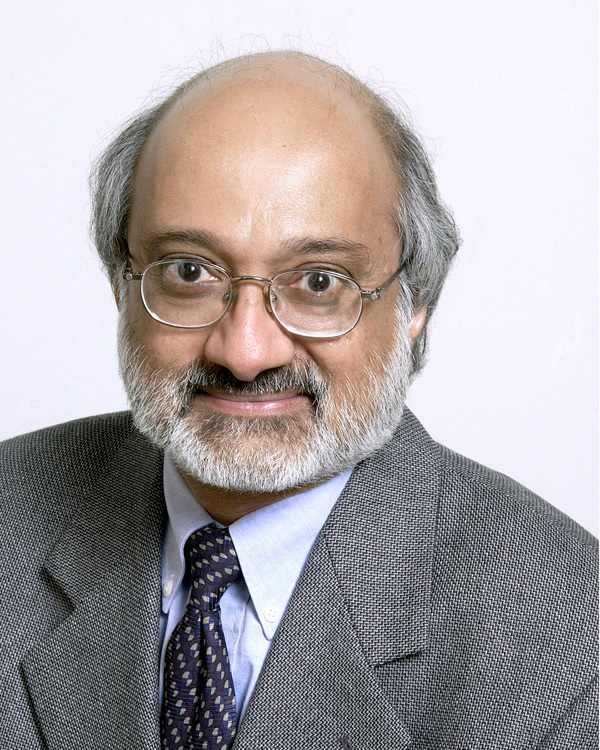}}]{Rajni V. Patel} (Life Fellow, IEEE) received the Ph.D. degree in electrical engineering from the University of Cambridge, Cambridge, U.K., in 1973. He is currently a Distinguished University Professor and the Tier-1 Canada Research Chair with the Department of Electrical and Computer Engineering, Western University, London, ON, Canada, with cross appointments in the School of Biomedical Engineering, the Department of Surgery and the Department of Clinical Neurological Sciences. He is also the Director of Engineering at Canadian Surgical Technologies and Advanced Robotics(CSTAR), Lawson Health Research Institute, London, ON, Canada. 
	
	Dr. Patel is a Fellow of the ASME and the Royal Society of Canada and the Canadian Academy of Engineering. He was on the editorial boards of IEEE TRANSACTIONS ON ROBOTICS, IEEE/ASME TRANSACTIONS ON MECHATRONICS, IEEE TRANSACTIONS ON AUTOMATIC CONTROL, Automatica, and Journal of Medical Robotics Research, and is currently on the editorial board of the International Journal of Medical Robotics and the Computer Assisted Surgery.
\end{IEEEbiography}

\end{document}